\documentclass{article}

\usepackage{microtype}
\usepackage{graphicx}
\usepackage{subfigure}
\usepackage{booktabs} % for professional tables

\usepackage{hyperref}
\usepackage{ntheorem}

\usepackage{amsfonts}
\usepackage{amsmath}
\usepackage{bbding}
\usepackage[capitalise]{cleveref}
\usepackage{mathrsfs}
\usepackage{dsfont}
\usepackage{amsmath,amsfonts,bm}

\def\eqref#1{equation~\ref{#1}}
\def\1{\bm{1}}

\DeclareMathAlphabet{\mathsfit}{\encodingdefault}{\sfdefault}{m}{sl}
\SetMathAlphabet{\mathsfit}{bold}{\encodingdefault}{\sfdefault}{bx}{n}

\newcommand{\Ls}{\mathcal{L}}
\newcommand{\R}{\mathbb{R}}

\DeclareMathOperator*{\argmin}{arg\,min}

\newcommand{\X}{\mathbf{X}} % "batch" of data subject features
\newcommand{\Y}{\mathbf{Y}} % "batch" of data subject labels
\newcommand{\xq}{X_\dagger} % query subject features
\newcommand{\yq}{Y_\dagger} % query subject true label
\newcommand{\yqt}{\tilde Y_\dagger} % query subject *target* prediction
\usepackage[accepted]{icml2021}

\icmltitlerunning{Online Algorithmic Recourse by Collective Action}

\begin{document}

\twocolumn[
\icmltitle{
Online Algorithmic Recourse by Collective Action%
}



\icmlsetsymbol{equal}{*}

\begin{icmlauthorlist}
\icmlauthor{Elliot Creager}{to,ve}
\icmlauthor{Richard Zemel}{to,ve}
\end{icmlauthorlist}

\icmlaffiliation{to}{University of Toronto}
\icmlaffiliation{ve}{Vector Institute}

\icmlcorrespondingauthor{Elliot Creager}{creager@cs.toronto.edu}

\icmlkeywords{Recourse, Collective Recourse, User Agency}

\vskip 0.3in
]



\printAffiliationsAndNotice{}  % leave blank if no need to mention equal contribution

\begin{abstract}
Research on algorithmic recourse typically considers how an individual can reasonably change an unfavorable automated decision when interacting with a fixed decision-making system.
This paper focuses instead on the online setting, where system parameters are updated dynamically according to interactions with data subjects.
Beyond the typical individual-level recourse, the online setting opens up new ways for groups to shape system decisions by leveraging the parameter update rule.
We show empirically that recourse can be improved when users coordinate by jointly computing their feature perturbations, underscoring the importance of collective action in mitigating adverse automated decisions.
\end{abstract}

\section{Introduction}
\newcommand{\nOne}{\mathcal{N}_1}
\newcommand{\nTwo}{\mathcal{N}_2}
\newcommand{\hOne}{\mathcal{H}_1}
\newcommand{\hTwo}{\mathcal{H}_2}
Pressing legal and ethical considerations have motivated a recent flurry of research on algorithmic recourse.
Generally, the technical framing of recourse computation involves a solving a constrained optimization problem that, given a fixed classifier $f_\theta$, finds the minimal amount of (plausible or actionable) effort required by individual $i$ to reverse an undesirable model prediction $\hat Y_i = f_\theta(X_i)$ given their features $X_i$ \citep{ustun2019actionable,karimi2020survey}.
In some applications, however, the model parameters $\theta$ may update upon deployment after new data are collected.
This opens up new avenues for realizing recourse, as the model behavior can be influenced indirectly through the parameter update rule.
Because training and test inputs are typically distinct, we will consider new types of recourse where (1) an individual seeks additional recourse assistance from \emph{others} whose data is used to train the model; and (2) those providing this additional recourse act \emph{collectively} towards this goal.

Consider a machine learning system that determines (or partially determines) high school assignment for an amalgamated school district comprising two neighborhoods $\{\nOne, \nTwo\}$ and two high schools $\{\hOne, \hTwo\}$.
The school district's model $f_\theta$ is trained to assign 8th graders to high schools based on a labeled dataset of input features $X_i$ (covariates collected about students prior to starting high school) and a label $Y_i$ indicating academic success after a period of time, such as graduation or GPA.
In principle, the model seeks to assign students to the school where their chances for success will be maximized.
In practice, it may learn a policy that (mostly) segregates students from $\nOne$ into $\hOne$, and from $\nTwo$ into $\hTwo$.
This could arise due to seemingly innocuous proxies in the training data, such as number of extracurricular activities, perhaps higher in $\nOne$ where more sports facilities are available, correlating with academic success in $\hOne$.

Now suppose that $\hOne$ offers more Advanced Placement (AP) courses than $\hTwo$.
Consider a subset of high-achieving students $S \subset \nTwo$ that want to take AP classes, but were assigned $f_\theta(X_i) = \hTwo \ \forall X_i \in S$.
While the ambition of these students suggests they would be well equipped to succeed in $\hOne$, they are nevertheless assigned to $\hTwo$ due to their low number of extracurricular activities.
What recourse is available to them to change their school assignment?

One way to compute (individualized) recourse for $i \in S$ is to assume a fixed $f_\theta$, then determine an actionable feature perturbation  $\delta$ (e.g. increasing extracurricular activities) such that $f_\theta(X_i+\delta_i) = \hOne$.
While this may prove useful the specific $i \in S$ in question, we focus on a different avenue for recourse that takes a longer-term perspective by leveraging the learning rule for the model parameters $\theta$.
If we assume that $\theta$ is updated each year using the graduating classes from $\{\hOne, \hTwo\}$ as a new training set, then all current students from $\hOne$ can collectively help prospective students from $S$ by under-reporting their extracurricular activity, causing $f_\theta$ to ignore this proxy variable when predicting next year's admissions cohort.

This method of \emph{collective recourse} is actionable, coordinated, and arguably provides a better long-term fix to the neighborhood diversity problem across both schools, all without relying on the district officials who deploy $f_\theta$ (who may or may not care about the long-term fairness of their system).
Crucially, this type of recourse requires that $f_\theta$ updates \emph{online} as new data becomes available.
To further investigate this issue, we present a technical framework for Online Algorithmic Recourse (OAR), and present preliminary evidence that collective action can be an effective way to shape model behavior in the online setting.
Our approach is inspired by recent advances in adversarial machine learning that also exploit knowledge of a learning rule to find small train-time perturbations that change test-time predictions.

%%%%%%%%%%%%%%%%%%%%%%%%%%%%%%%%%%%%%%%%%%%%%%%%%%%%%%%%%%%%
% fig 1
%%%%%%%%%%%%%%%%%%%%%%%%%%%%%%%%%%%%%%%%%%%%%%%%%%%%%%%%%%%%
\begin{figure*}[t!]
\centering
\subfigure{
\includegraphics[width=.8\columnwidth]{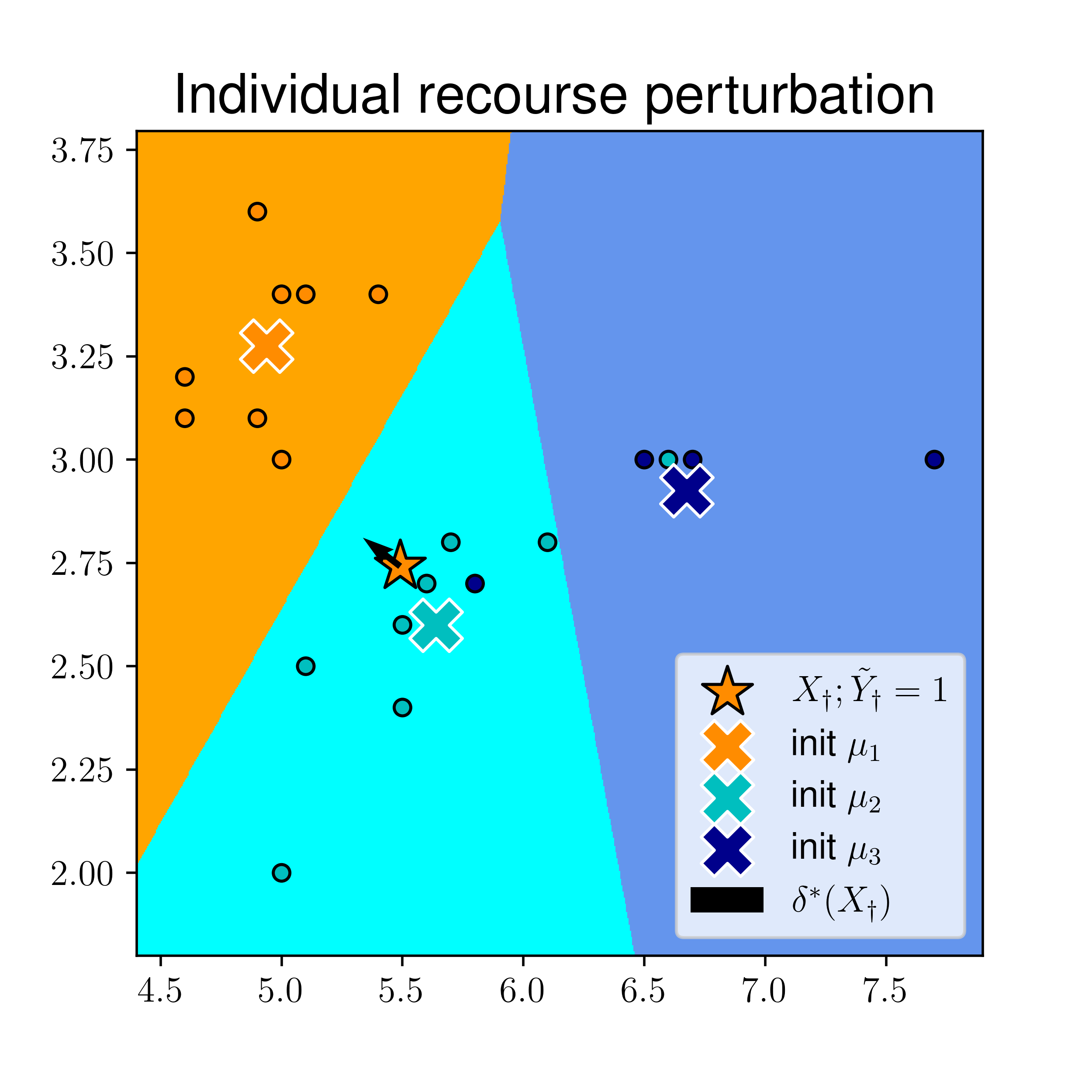}
}
% \hfill
\subfigure{
\includegraphics[width=.8\columnwidth]{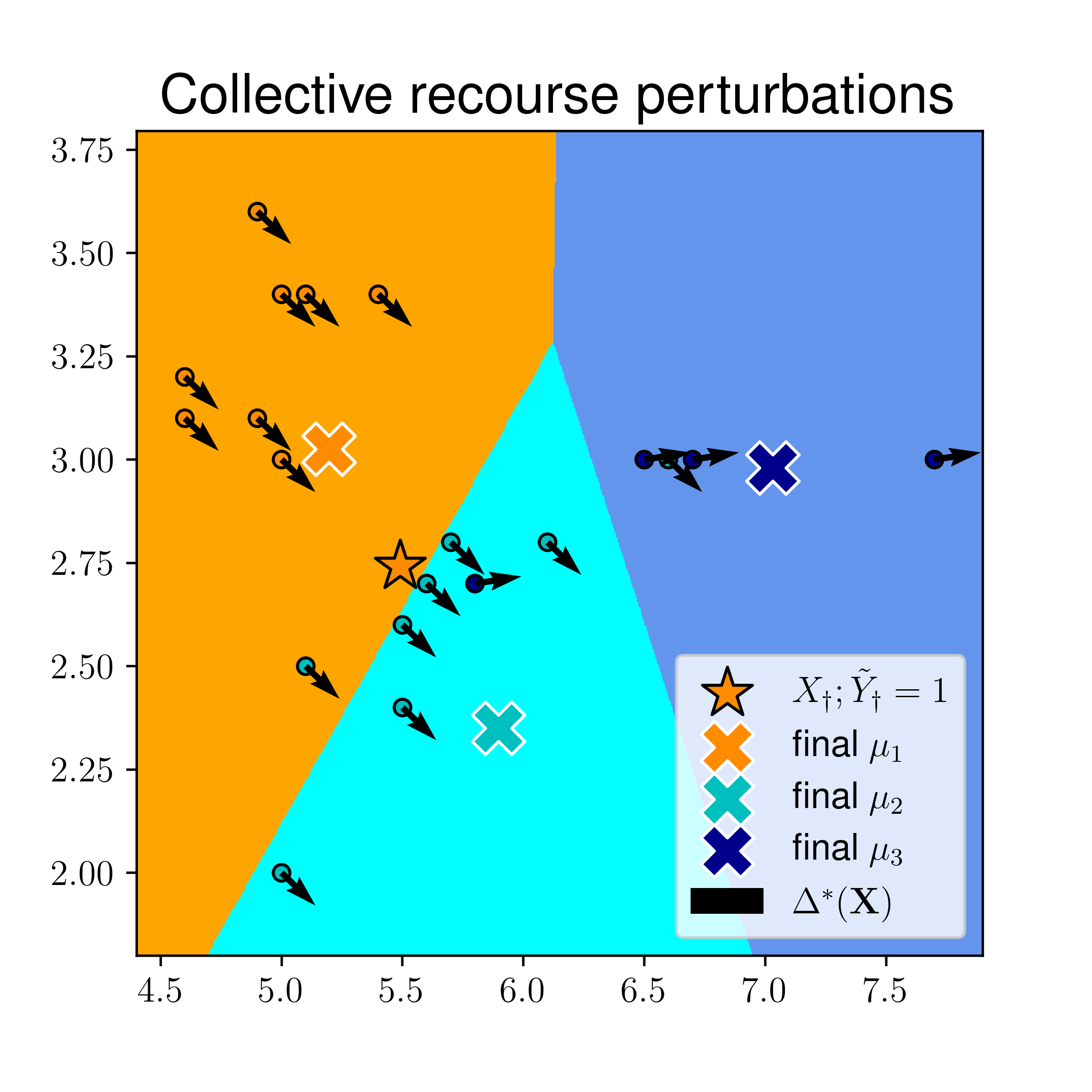}
}
\caption[]{
Online algorithmic recourse under a nearest-centroid classifier.
\textbf{Left:}
A nearest-centroid classifier with parameters ${\theta = \{\mu_1, \mu_2, \mu_3\}}$ is fit to 20 \emph{data subjects}, with features $\X$ (circles) and class labels $\Y$ (circle colors).
A \emph{query subject} $\xq$ seeks to change their prediction under this model to $\yqt = 1$.
The query subject can exercise \emph{individual recourse} by applying the best perturbation $\delta$ with $||\delta|| \leq \epsilon$ towards this end, moving $\xq+\delta$ towards the decision boundary separating $\hat Y = 1$ from $\hat Y = 2$.
\textbf{Right:}
Another means of recourse is to have \emph{all} data subjects coordinate by applying perturbations $\Delta = [\delta_1 \ldots \delta_N]^T$ with ${||\delta_i|| \leq \epsilon \ \forall \ i}$ in order to help the query subject.
Note that this changes the model's decision boundary, as the centroids $\{\mu_k\}$ are fit to the perturbed data $(\X+\Delta, \Y)$.
This \emph{collective recourse} is more effective in this instance, as we see that once the model fits new centroids to the perturbed training data, $\xq$ lies on the desired side of the model decision boundary.
Data in this example come from first two features of the UCI-Iris dataset.
}
\label{fig:fig1}
\end{figure*}
%%%%%%%%%%%%%%%%%%%%%%%%%%%%%%%%%%%%%%%%%%%%%%%%%%%%%%%%%%%%

\section{Method}

\paragraph{Preliminaries}
Denote by $X_i \in \R^{d}$ an observation of $d$ features from the $i$-th ``data subject'' (i.e. individual), and by $Y_i \in \R^{k}$ its associated target, which could be a one-hot encoded class label for $k$-way classification, or real-valued vector for $k$-dimensional regression.
In some cases $Y_i$ may correspond to a real-world outcome, as in the school admissions case mentioned above; other times they may be assigned by a labeler (often a crowd worker).
When observations from $N$ individuals are collected at once, they make a ``batch'' of labeled data $(\X, \Y)$ with $\X \in \R^{N \times d}$ and $\Y \in \R^{N \times k}$, i.e. $\X = [X_1 \ldots X_N]^T$.
The model is specified by a prediction function $f_\theta: \R^d \rightarrow \R^k$ with parameters $\theta$, along with a scalar loss function $\Ls: \R^k \times \R^k \rightarrow \R$ that scores a model prediction $\hat y = f_\theta(x)$ w.r.t its target $y$.
Crucially, the designers of the model assume the individuals within $\X$ to be \emph{distributed i.i.d.}, $p(\X) = \prod_i p(X_i)$, which justifies learning the parameters by minimizing the empirical risk $R(\theta, \X, \Y)$ as follows:

\begin{align}
    \theta^* = \argmin_\theta 
    \underbrace{
    \frac{1}{N} \sum_{i=1}^{N} \Ls(f_\theta(X_i), Y_i). \label{eq:erm}
    }_{
    R(\theta, \X, \Y)
    }
\end{align}
Any suitable regularizer(s), e.g. weight decay, can be added onto the empirical risk $R(\theta, \X, \Y)$ as needed, but we omit them here for ease of exposition.

This paper is concerned with measuring and improving a particular type of recourse, where the data subjects comprising the ``batch'' $\X$ seek to change the model's behavior when making predictions or decisions about some \emph{other}  individual(s), the \emph{query subject(s)}.
In general, there may be multiple query subjects, and the data subject and query subject groups may be overlapping, but for simplicity of exposition, we consider only one such \emph{query subject}, and denote its features and (true) target as $(\xq, \yq)$.
Regardless of the true value of the query target $\yq$, the query subject seeks to have the model predict a particular \emph{goal target} $\yqt$ on the query subjects features, i.e. recourse is successful if $f_\theta(\xq) = \yqt$.
As we discuss next, this can be accomplished by the query subject perturbing their features (individual recourse), or by a coordinated effort of the data subjects on behalf of the query subject (collective recourse).

\paragraph{Individual recourse}
Given the query subject's desire to set their model prediction to $\yqt$, the most immediate avenue for recourse is for the query subject to directly perturb their inputs.
Assuming that such a perturbation must have a small $p$-norm\footnote{
We measure size of the perturbation using a $p$-norm here for convenience, but this could easily be replaced with another suitable metric, e.g. indicating a cost or feasibility of recourse per feature \citep{ustun2019actionable}.
} of $\epsilon \geq 0$, this means of \emph{individual recourse} can be summarized by the following optimization problem:
\begin{align}
    \delta^*(\xq) &= \argmin_\delta \Ls(f(\xq + \delta), \yqt) \nonumber\\
    \text{s.t.} \quad ||\delta||_p & \leq \epsilon.
    \label{eq:ir}
\end{align}
By encouraging the model assign low loss for the perturbed input when $Y = \yqt$, the query subject maximizes the chances that the model will predict $\hat Y = \yqt$ on the perturbed $\xq$.

\paragraph{Online algorithmic recourse}
We now examine the leverage that data subjects have on the query subject's model prediction.
If $f_\theta$ is fixed and learned ahead of time (i.e. the data subjects are not part of the training data), then its predictions will be rendered independently, and the data subjects have no influence on the fate of the query subject.\footnote{
One subtle exception is when $f_\theta$ is a deep neural net trained with batch norm, where it is somewhat common for the batch norm parameters to be updated at test time by computing activations statistics over the test inputs (test labels are not required for this). 
Thus it is possible that data subjects could affect the query subject prediction by adversarially attacking the batch norm parameters at test time.
}
However, if the model updates its parameters \emph{online}, the data subjects can realize leverage over model behavior through its update rule.

Assuming an online regime, we allow each data subject $X_i$ to make a small perturbation $\delta_i$ on its features, where $||\delta_i||_p \leq \epsilon$.
The data subjects will assert recourse on behalf of the query subject by affecting the model predictions via the model parameter updates from \cref{eq:erm}; specifically, they try to encourage the model to have low loss on $(\xq, \yqt)$, i.e. $f_\theta(\xq) \approx \yqt$.
Letting $\Delta = [\delta_1 \ldots \delta_N]^T$ denote the batched perturbations for all data subjects and $\text{rows}(\Delta)$ the set of its unique row vectors, this implies the following bi-level optimization problem:
\begin{align}
    \Delta^* &= \argmin_{\Delta} \Ls(f_{\theta^*(\Delta)}(\xq), \yqt) \nonumber\\
    \text{s.t.} \quad \theta^*(\Delta) &= \argmin_\theta R(\theta, \X + \Delta, \Y) \nonumber \\
    ||\delta_i||_p & \leq \epsilon \ \forall \ \delta_i \in \text{rows}(\Delta).
    \label{eq:oar}
\end{align}
Because the data subjects exercise recourse through the model's online parameter updates, we dub this Online Algorithmic Recourse (OAR).
We note that this framework can be extended to handle multiple timesteps of model updates, which we leave for future work.

\paragraph{Collective action}
A solution to \cref{eq:oar} can be seen as a sort of \emph{collective recourse}, where the data subjects act to change the model behavior on behalf of the query subject.
We verify in the experiments below that, in a simple setting where good solutions to \cref{eq:oar} can be readily computed, this collective action can actually be \emph{more effective} than if the query subject directly perturbs their features as in \cref{eq:ir}.
Recall the key assumption underpinning the model design: that training and test points are drawn i.i.d. from the same distribution.
By coordinating their actions on behalf of the query subject, the data subjects \emph{violate the i.i.d. assumption by collective action}, thus realizing considerable leverage over model behavior.
OAR provides a simple and intuitive framework in which to explore this theme.
\cref{fig:fig1} illustrates the difference between collective and individual recourse.

\section{Related work}
A key inspiration for the OAR framework is the recent work on computing adversarial training examples, often referred to as Data Poisoning (DP) \cite{koh2017understanding,shafahi2018poison,shan2020fawkes,geiping2020witches,cherepanova2021lowkey}.
Generally, DP seeks to degrade test-time predictive performance via imperceptible perturbations to (some of) the training set.
For example, \citet{geiping2020witches} used bi-level constrained optimization (similar to OAR) to poison a subset of training data using a gradient alignment procedure on an auxiliary classifier trained from public data. 
In some instances OAR and DP may be strongly aligned, e.g. when a nefarious online surveillance tool is deployed and the data subjects seek recourse by reducing overall model accuracy with imperceptible input perturbations \citep{albert2020politics}.
In other instances, OAR may differ by seeking perceptible (but actionable) perturbations that change model behavior in more targeted way, i.e. for a specific query subject rather than the global population.

Our work is also inspired by Protective Optimization Technologies (POTs) \citep{kulynych2020pots}, a broad framework that discusses the countermeasures available to data subjects facing automated decisions rendered by an untrustworthy party.
For example \citet{kulynych2020pots} present a greedy algorithm for selectively curating a training set to ensure particular fairness criteria in $f_\theta$ are realized; by contrast, our empirical approach focuses on additive perturbations on the features, which are optimized jointly via gradient-based methods.

OAR also bears a resemblance to strategic classification \citep{hardt2016strategic,milli2019social}, since data subjects change their features to encourage certain model behaviors.
One difference is that whereas strategic classifiers seek equilibrium in a Stackleberg game where the model leads and the data subjects follow, in our setup the roles are reversed.
\citet{perdomo2020performative} proposed Performative Prediction (PP), whereby the data distribution changes as a function of the model parameters $p(\X,\Y) = \mathcal{D}(\theta)$, and a model stability w.r.t. this dynamic is sought.
Whereas PP assumes a fixed mapping $\mathcal{D}$, OAR can be phrased as computing an optimal mapping $\mathcal{D}$ from the perspective of maximizing recourse.

The Algorithmic Recourse literature has recently taken steps towards moving ``beyond individualized recourse'' \citep{rawal2020beyond, karimi2020survey}, by computing population-level recommendations \citep{rawal2020beyond}, or examining situations where one data subject seeks actionable recourse on behalf of another \citep{venkatasubramanian2020philosophical}.
Our approach is distinct in its focus on the online setting, specifically examining recourse that data subjects have through the update to $\theta$.

\section{Experiments}
In this section we present a simple proof-of-concept experiment showing that recourse can be improved by viewing the problem through an online lens, then solving for the optimal collective (rather than individual) action.

\paragraph{Setup}
We use the following simple classification datasets:
\begin{itemize}
    \item UCI-Iris:\footnote{\mbox{\scriptsize \url{https://archive.ics.uci.edu/ml/datasets/iris}}} we use $(\X, \Y)$ as the $N=150$ examples of $d=4$ attributes are collected from each of $k=3$ flower types;
    \item MNIST:\footnote{\mbox{\scriptsize    \url{http://yann.lecun.com/exdb/mnist/}}} we use $(\X, \Y)$ as the $N=60,000$ handwritten digits stored as $28 \times 28$-pixel greyscale images, with $k=10$ classes corresponding digits 0--9.
    Prior to computing recourse we embed the images in a pre-trained VAE embedding space with $d=10$ dimensions.
\end{itemize}
For each dataset, we compute individual recourse by solving \cref{eq:ir} and collective recourse by solving \cref{eq:oar}, using $p=2$ to compute $p$-norms in the constraint.
We choose a synthetic query point $\xq$ that lies soundly within the model's decision boundary for $\hat Y = f_\theta(\xq) = 2$, then specify $\yqt = 1$, meaning that the query prediction should be flipped.
For lack of space, we defer details to \cref{sec:experimental-details}.

%%%%%%%%%%%%%%%%%%%%%%%%%%%%%%%%%%%%%%%%%%%%%%%%%%%%%%%%%%%%
% fig 2
%%%%%%%%%%%%%%%%%%%%%%%%%%%%%%%%%%%%%%%%%%%%%%%%%%%%%%%%%%%%
\begin{figure}[h]
\centering
\subfigure{
\includegraphics[width=.22\textwidth]{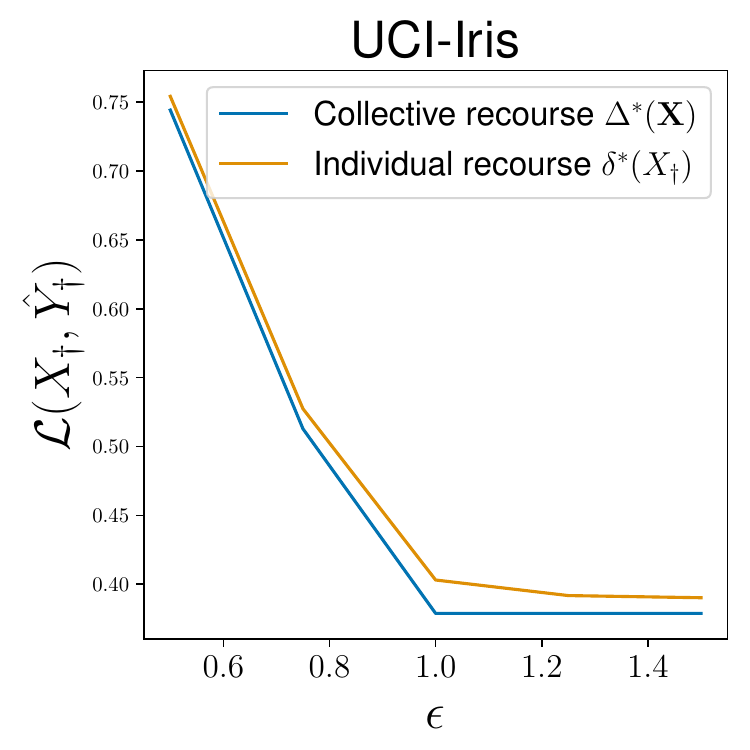}
}
\hfill
\subfigure{
\includegraphics[width=.22\textwidth]{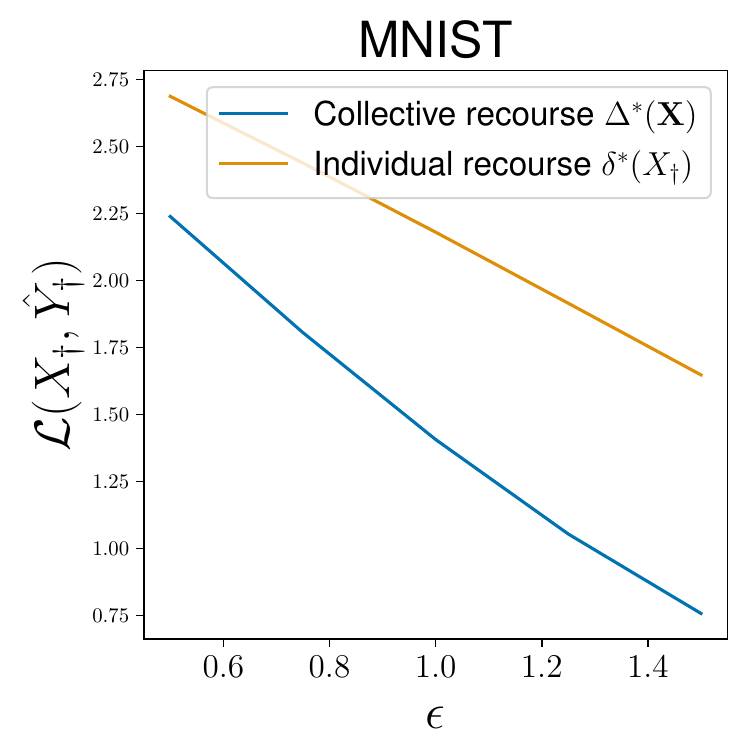}
}
\caption[]{
Measuring success of recourse according to model loss for query goal class $\yqt$ evaluated at query input $\xq$ (the lower the better).
The model is a nearest centroid classifier.
We find that on UCI-Iris (left), and MNIST (right), collective recourse is more effective than individual recourse at a given $\epsilon$ budget.
}
\label{fig:fig2}
\end{figure}
%%%%%%%%%%%%%%%%%%%%%%%%%%%%%%%%%%%%%%%%%%%%%%%%%%%%%%%%%%%%

\paragraph{Results}
While the data subjects have only an indirect influence on the model through its parameter update, we find that this influence is substantial in practice.
In fact, across all $\epsilon$ budgets, we find that collective recourse is more effective than individual recourse, as we see that for a given $\epsilon$, the loss $\Ls(\xq, \yqt)$ is lower (as desired) for the collective perturbations than the individual perturbations (\cref{fig:fig2}). 
In other words, the perturbations on the data subjects' inputs $\X$, which were computed by phrasing recourse as an \emph{online} problem, change the model behavior for the query subject more than if the query subject acts on their own behalf to directly perturb their inputs $\xq$.

\section{Discussion}
In this paper we drew inspiration from the adversarial machine learning literature to present the framework of Online Algorithmic Recourse (OAR), and presented preliminary evidence that there may be power in numbers when it comes to achieving recourse in this setting.
That is, data subjects can coordinate to mitigate undesirable prediction on separate query subject(s) because their data determines the model's update rule. 
However there remains much to be done in this area.
For example, while we constrained the perturbations by their $p$-norm for convenience, alternative constraints that represent actionability, feasibility, or diversity of recourse could enable the adoption of techniques from the counterfactual explanation literature \citep{ustun2019actionable,poyiadzi2020face,dandl2020multi} to the computation of collective recourse.
Also, our experiments assumed the data subjects have access to the model's loss function (analogous to white-box adversarial attacks).
In the future we would like to relax this assumption, noting that adversarial attacks often generalize across architectures \citep{demontis2019why}, so that black-box attacks can be realized by training auxiliary models using public data \citep{geiping2020witches}.
Finally, we have thusfar ignored the role of the labeling process.
When labels are acquired using crowd workers, the labelers themselves have plenty of leverage over model behavior that could be used for algorithmic recourse, and are already known to work collaboratively \citep{irani2013turkopticon}.

\bibliography{refs}

\begin{thebibliography}{22}
\providecommand{\natexlab}[1]{#1}
\providecommand{\url}[1]{\texttt{#1}}
\expandafter\ifx\csname urlstyle\endcsname\relax
  \providecommand{\doi}[1]{doi: #1}\else
  \providecommand{\doi}{doi: \begingroup \urlstyle{rm}\Url}\fi

\bibitem[Albert et~al.(2020)Albert, Penney, Schneier, and
  Siva~Kumar]{albert2020politics}
Albert, K., Penney, J., Schneier, B., and Siva~Kumar, R.~S.
\newblock Politics of adversarial machine learning.
\newblock In \emph{Towards Trustworthy ML: Rethinking Security and Privacy for
  ML Workshop, Eighth International Conference on Learning Representations
  (ICLR)}, 2020.

\bibitem[Cherepanova et~al.(2021)Cherepanova, Goldblum, Foley, Duan, Dickerson,
  Taylor, and Goldstein]{cherepanova2021lowkey}
Cherepanova, V., Goldblum, M., Foley, H., Duan, S., Dickerson, J., Taylor, G.,
  and Goldstein, T.
\newblock Lowkey: leveraging adversarial attacks to protect social media users
  from facial recognition.
\newblock 2021.

\bibitem[Dandl et~al.(2020)Dandl, Molnar, Binder, and Bischl]{dandl2020multi}
Dandl, S., Molnar, C., Binder, M., and Bischl, B.
\newblock Multi-objective counterfactual explanations.
\newblock In \emph{International Conference on Parallel Problem Solving from
  Nature}, pp.\  448--469. Springer, 2020.

\bibitem[Demontis et~al.(2019)Demontis, Melis, Pintor, Jagielski, Biggio,
  Oprea, Nita-Rotaru, and Roli]{demontis2019why}
Demontis, A., Melis, M., Pintor, M., Jagielski, M., Biggio, B., Oprea, A.,
  Nita-Rotaru, C., and Roli, F.
\newblock Why do adversarial attacks transfer? explaining transferability of
  evasion and poisoning attacks.
\newblock In \emph{Proceedings of the 28th USENIX Conference on Security
  Symposium}, SEC'19, pp.\  321–338, USA, 2019. USENIX Association.
\newblock ISBN 9781939133069.

\bibitem[Geiping et~al.(2021)Geiping, Fowl, Huang, Czaja, Taylor, Moeller, and
  Goldstein]{geiping2020witches}
Geiping, J., Fowl, L., Huang, W.~R., Czaja, W., Taylor, G., Moeller, M., and
  Goldstein, T.
\newblock Witches' brew: Industrial scale data poisoning via gradient matching.
\newblock In \emph{International Conference on Learning Representations}, 2021.

\bibitem[Goodfellow et~al.(2014)Goodfellow, Pouget-Abadie, Mirza, Xu,
  Warde-Farley, Ozair, Courville, and Bengio]{goodfellow2014generative}
Goodfellow, I., Pouget-Abadie, J., Mirza, M., Xu, B., Warde-Farley, D., Ozair,
  S., Courville, A., and Bengio, Y.
\newblock Generative adversarial nets.
\newblock In Ghahramani, Z., Welling, M., Cortes, C., Lawrence, N., and
  Weinberger, K.~Q. (eds.), \emph{Advances in Neural Information Processing
  Systems}, volume~27. Curran Associates, Inc., 2014.
\newblock URL
  \url{https://proceedings.neurips.cc/paper/2014/file/5ca3e9b122f61f8f06494c97b1afccf3-Paper.pdf}.

\bibitem[Hardt et~al.(2016)Hardt, Megiddo, Papadimitriou, and
  Wootters]{hardt2016strategic}
Hardt, M., Megiddo, N., Papadimitriou, C., and Wootters, M.
\newblock Strategic classification.
\newblock In \emph{Proceedings of the 2016 ACM conference on innovations in
  theoretical computer science}, pp.\  111--122, 2016.

\bibitem[Irani \& Silberman(2013)Irani and Silberman]{irani2013turkopticon}
Irani, L.~C. and Silberman, M.~S.
\newblock Turkopticon: Interrupting worker invisibility in amazon mechanical
  turk.
\newblock In \emph{Proceedings of the SIGCHI conference on human factors in
  computing systems}, pp.\  611--620, 2013.

\bibitem[Karimi et~al.(2020)Karimi, Barthe, Sch{\"o}lkopf, and
  Valera]{karimi2020survey}
Karimi, A.-H., Barthe, G., Sch{\"o}lkopf, B., and Valera, I.
\newblock A survey of algorithmic recourse: definitions, formulations,
  solutions, and prospects.
\newblock \emph{arXiv preprint arXiv:2010.04050}, 2020.

\bibitem[Koh \& Liang(2017)Koh and Liang]{koh2017understanding}
Koh, P.~W. and Liang, P.
\newblock Understanding black-box predictions via influence functions.
\newblock In \emph{International Conference on Machine Learning}, pp.\
  1885--1894. PMLR, 2017.

\bibitem[Kulynych et~al.(2020)Kulynych, Overdorf, Troncoso, and
  G{\"u}rses]{kulynych2020pots}
Kulynych, B., Overdorf, R., Troncoso, C., and G{\"u}rses, S.
\newblock Pots: protective optimization technologies.
\newblock In \emph{Proceedings of the 2020 Conference on Fairness,
  Accountability, and Transparency}, pp.\  177--188, 2020.

\bibitem[Lorraine \& Duvenaud(2018)Lorraine and
  Duvenaud]{lorraine2018stochastic}
Lorraine, J. and Duvenaud, D.
\newblock Stochastic hyperparameter optimization through hypernetworks.
\newblock \emph{arXiv preprint arXiv:1802.09419}, 2018.

\bibitem[Maclaurin et~al.(2015)Maclaurin, Duvenaud, and
  Adams]{maclaurin15gradient}
Maclaurin, D., Duvenaud, D., and Adams, R.
\newblock Gradient-based hyperparameter optimization through reversible
  learning.
\newblock In Bach, F. and Blei, D. (eds.), \emph{Proceedings of the 32nd
  International Conference on Machine Learning}, volume~37 of \emph{Proceedings
  of Machine Learning Research}, pp.\  2113--2122, Lille, France, 07--09 Jul
  2015. PMLR.
\newblock URL \url{http://proceedings.mlr.press/v37/maclaurin15.html}.

\bibitem[Milli et~al.(2019)Milli, Miller, Dragan, and Hardt]{milli2019social}
Milli, S., Miller, J., Dragan, A.~D., and Hardt, M.
\newblock The social cost of strategic classification.
\newblock In \emph{Proceedings of the Conference on Fairness, Accountability,
  and Transparency}, FAT* '19, pp.\  230–239, New York, NY, USA, 2019.
  Association for Computing Machinery.
\newblock ISBN 9781450361255.
\newblock \doi{10.1145/3287560.3287576}.
\newblock URL \url{https://doi.org/10.1145/3287560.3287576}.

\bibitem[Perdomo et~al.(2020)Perdomo, Zrnic, Mendler-D{\"u}nner, and
  Hardt]{perdomo2020performative}
Perdomo, J., Zrnic, T., Mendler-D{\"u}nner, C., and Hardt, M.
\newblock Performative prediction.
\newblock In \emph{International Conference on Machine Learning}, pp.\
  7599--7609. PMLR, 2020.

\bibitem[Poyiadzi et~al.(2020)Poyiadzi, Sokol, Santos-Rodriguez, De~Bie, and
  Flach]{poyiadzi2020face}
Poyiadzi, R., Sokol, K., Santos-Rodriguez, R., De~Bie, T., and Flach, P.
\newblock Face: Feasible and actionable counterfactual explanations.
\newblock In \emph{Proceedings of the AAAI/ACM Conference on AI, Ethics, and
  Society}, pp.\  344--350, 2020.

\bibitem[Rawal \& Lakkaraju(2020)Rawal and Lakkaraju]{rawal2020beyond}
Rawal, K. and Lakkaraju, H.
\newblock Beyond individualized recourse: Interpretable and interactive
  summaries of actionable recourses.
\newblock \emph{Advances in Neural Information Processing Systems}, 33, 2020.

\bibitem[Shafahi et~al.(2018)Shafahi, Huang, Najibi, Suciu, Studer, Dumitras,
  and Goldstein]{shafahi2018poison}
Shafahi, A., Huang, W.~R., Najibi, M., Suciu, O., Studer, C., Dumitras, T., and
  Goldstein, T.
\newblock Poison frogs! targeted clean-label poisoning attacks on neural
  networks.
\newblock In \emph{Neural Information Processing Systems}, 2018.

\bibitem[Shan et~al.(2020)Shan, Wenger, Zhang, Li, Zheng, and
  Zhao]{shan2020fawkes}
Shan, S., Wenger, E., Zhang, J., Li, H., Zheng, H., and Zhao, B.~Y.
\newblock Fawkes: Protecting privacy against unauthorized deep learning models.
\newblock In \emph{29th $\{$USENIX$\}$ Security Symposium ($\{$USENIX$\}$
  Security 20)}, pp.\  1589--1604, 2020.

\bibitem[Snell et~al.(2017)Snell, Swersky, and Zemel]{snell2017prototypical}
Snell, J., Swersky, K., and Zemel, R.~S.
\newblock Prototypical networks for few-shot learning.
\newblock In \emph{Neural Information Processing Systems}, 2017.

\bibitem[Ustun et~al.(2019)Ustun, Spangher, and Liu]{ustun2019actionable}
Ustun, B., Spangher, A., and Liu, Y.
\newblock Actionable recourse in linear classification.
\newblock In \emph{Proceedings of the Conference on Fairness, Accountability,
  and Transparency}, pp.\  10--19, 2019.

\bibitem[Venkatasubramanian \& Alfano(2020)Venkatasubramanian and
  Alfano]{venkatasubramanian2020philosophical}
Venkatasubramanian, S. and Alfano, M.
\newblock The philosophical basis of algorithmic recourse.
\newblock In \emph{Proceedings of the 2020 Conference on Fairness,
  Accountability, and Transparency}, FAT* '20, pp.\  284–293, New York, NY,
  USA, 2020. Association for Computing Machinery.
\newblock ISBN 9781450369367.
\newblock \doi{10.1145/3351095.3372876}.
\newblock URL \url{https://doi.org/10.1145/3351095.3372876}.

\end{thebibliography}
\bibliographystyle{icml2021}

\appendix

\section{Experimental details}\label{sec:experimental-details}
For each dataset, we fit a nearest centroid classifier to $(\X, \Y)$.
This is a distance-based classifier that assigns predictive probability of $X$ belonging to class $Y=y$ as 
\[
p(Y=y|X) = \frac{\exp(-||X - \mu_y||)}{\sum_{y'=1}^{k}\exp(-||X - \mu_{y'}||)}.
\]
The model parameters are the per-class centroid vectors ${\theta = \{\mu_y\} \ \forall \ y \in 1..k}$, which are fit by maximum likelihood (equivalent to computing per-class averages in the data space).

We note that, despite their simplicity, distance-based classifiers such as nearest centroid are fairly common in deep learning (applied within a learned embedding space), especially in few-shot learning \citep{snell2017prototypical}.
The base model achieves training accuracies of 96\% and 83\% on UCI-Iris and MNIST, respectively.

For the MNIST dataset, rather than fitting centroids to the pixels directly, we first embedded the images into a learned $10$-dimensional VAE embedding space.
For the VAE encoder and decoder, we used two-hidden-layer MLPs with $512$ hidden neurons per layer, which were trained for $500$ epochs using SGD with learning rate $0.001$ and momentum $0.9$.
Following training, we embed the images by directly using the encoder mean for each image, rather than sampling from a Gaussian with the encoder mean and variance as parameters.

We choose $f_\theta$ to be a nearest-centroid classifier because the optimal post-update model parameters under perturbation, i.e. $\theta^*(\Delta)$ from \cref{eq:oar}, can be computed in closed form.
Thus substituting this solution into \cref{eq:oar} gives a single-level constrained optimization problem, instead of the bilevel problem required for generic $f_\theta$.
We note that several approaches to solving the generic bilevel optimization problem---e.g. unrolling or approximating the lower-level solution as in hyperparemeter optimization \citep{maclaurin15gradient,lorraine2018stochastic} or alternating updates as in GANs \citep{goodfellow2014generative}---could be employed here to tackle generic (twice) differentiable models $f_\theta$, but we leave this for future work.

We specify a query point as $\xq = \alpha \mu_1 + (1 - \alpha) \mu_2$ (we use $\alpha = 0.25$ in our experiments), with corresponding query goal target $\yqt = 1$.
The query input $\xq$ thus lies close to the $Y=2$ centroid, but its model prediction must be flipped so that $\yqt=1$.

We compute individual and collective recourse by approximately solving \cref{eq:ir} and \cref{eq:oar}, respectively.
For ease of implementation, we use a heuristic projected-gradient attack to solve each optimization problem.
Noting that, as mentioned above, the choice of nearest centroid model makes both \cref{eq:ir} and \cref{eq:oar} constrained optimization problems, we compute a gradient w.r.t the perturbation of the objective, then normalize the perturbation to have the desired norm of $\epsilon$.
\end{document}